\documentclass{article}

\usepackage[final, nonatbib]{new_in_ML}

\usepackage[
]{biblatex}

\usepackage[utf8]{inputenc} 
\usepackage[T1]{fontenc}    
\usepackage{hyperref}       
\usepackage{url}            
\usepackage{booktabs}       
\usepackage{amsfonts}       
\usepackage{nicefrac}       
\usepackage{microtype}      
\usepackage{xcolor}         
\usepackage{multicol}
\usepackage{graphicx}
\usepackage{capt-of}
\usepackage{floatrow}
\newfloatcommand{capbtabbox}{table}[][\FBwidth]

\title{A Novel Explanation Against Linear Neural Networks}

\author{Anish Lakkapragada $^{1}$ \\
$^1$Lynbrook High School, San Jose, CA 95129 \\
\texttt{\{Email:alakkapragada176@student.fuhsd.org\}}}

\begin{document}

\maketitle


\begin{abstract}
Linear Regression and neural networks are widely used to model data. Neural networks distinguish themselves from linear regression with their use of activation functions that enable modeling nonlinear functions. The standard argument for these activation functions is that without them, neural networks only can model a line. However, a novel explanation we propose in this paper for the impracticality of neural networks without activation functions, or linear neural networks, is that they actually reduce both training and testing performance. Having more parameters makes LNNs harder to optimize, and thus they require more training iterations than linear regression to even potentially converge to the optimal solution. We prove this hypothesis through an analysis of the optimization of an LNN and rigorous testing comparing the performance between both LNNs and linear regression on synthethic, noisy datasets.
\end{abstract}

\section{Introduction}

Neural networks \cite{mcculloch1943logical}  distinguish themselves from linear regression by their ability to model nonlinear data. This capability comes from their nonlinear activation functions. The standard explanation against neural networks without such activation functions, which we refer to as linear neural networks (LNNs), is that they only can model lines and thus yield no benefit compared to linear regression.

In this paper, we propose a novel reason for the impracticality of LNNs: LNNs  actually perform worse than linear regression, despite modeling the same form of data. The excess of parameters in LNNs corrupts the optimization process thus preventing LNN training to yield the optimal solution. We test our hypothesis through a debrief of optimization procedures on an LNN and perform experiments on synthethic datasets of various noisiness.

\section{Methods}

If we have a univariate dataset $X$ and associated labels $y$, assuming the relationship between $X$ and $y$ is linear, a linear regression model given by the equation $\hat{y}_{i}$ = $ax_{i} + b$ can be created where $\hat{y}_{i}$ is the prediction for the input $x_{i}$. If this model was fully optimized, $a$ and $b$ would be the weight and bias respectively to minimize the mean of the squared residuals. 

Neural networks for univariate data can similarly be constructed as the following. The output vector for the first layer $z_{1}$ is given by $z_{1} = w_{1}x + b_{1}$. $w_{n}$ and $b_{n}$ denote the weight and bias for the $n$th layer. The output of an LNN with a second layer would then be $w_{2}z_{1} + b_{2}$ or $w_{2}w_{1}x + w_{2}b_{1} + b_{2}$. 

LNNs require iterative optimization, such as Gradient Descent (GD), to optimally adjust their parameters. GD updates each of current
parameters based on the derivative of the objective function $j$
with respect to that parameter.Given learning rate $\alpha$ and any parameter at time step $t$, GD will update the parameter to $p_{t + 1}$ as such: $p_{t+1} = p_{t} - \alpha \frac{dJ}{dp}$. In our case, our objective function is the mean squared error (MSE) given by $J = \frac{1}{N} \sum_{i=1}^{N} (\hat{y}_{i} - y_{i})^{2}$. The derivatives used to optimize a linear regression parameters ${m, b}$ through such optimization are shown in Equation 1.

\begin{equation}
    \frac{dJ}{dm} = \frac{2}{N} \sum_{i=1}^{N} (\hat{y}_{i} - y_{i})x_{i}; \frac{dJ}{db} = \frac{2}{N} \sum_{i=1}^{N}(\hat{y}_{i} - y_{i})
\end{equation}


LNN optimization is more cumbersome because of the increased amount of parameters. For the two-layered LNN given by $w_{2}w_{1}x + w_{2}b_{1} + b_{2}$, the optimal parameter solution is for $w_{2}w_{1} = a; w_{2}b_{1} + b_{2} =  b$ so that the LNN's prediction function simplifies to the $ax + b$. Because the derivative of any parameter depends on parameters from previous layers, this makes this solution harder to reach. Given the derivative of $J$ with respect to $w_{2}$ used to optimize $w_{2}$: 

\begin{equation}
\frac{dJ}{dw_{2}} = \frac{2}{N} \sum_{i=1}^{N} (\hat{y}_{i} - y_{i}) (w_{1}x_{i} + b_{1})
\end{equation}

we can see that the next step of $w_{2}$ by GD would be based on the currently suboptimal parameters $w_{1}$ and $b_{1}$. In order for the optimal solution $w_{2}w_{1} = a$ to be met, this means the new value of $w_{2}$, calculated on a suboptimal $w_{1}$, and $w_{1}$ have to align such that their product is $a$. This will realistically only happen if the LNN begins training with a parametrization initialization where $w_{2}w_{1} = a$. GD initializes parameters randomly, so this particular arrangement is extremely unlikely. The high interdependency between parameters and their movements across iterations creates difficulty for an LNN's parameters to arrive at the optimal solution. Note that these same dynamics apply to the optimization of the bias parameter. Through this demonstration, it can be seen how this problem will be further exacerbated if the LNN had more layers, and thus more parameters.



\section{Experiments}

We compare the performance of linear regression and LNNs from 2 to 10 layers on synthetic datasets with varying levels of noise.

\subparagraph{Data}
For simplicity, all of our data in our experiments are univariate. Note that even if our data was multivariate, the same results would occur as linear regression or LNNs on multivariate data essentially operates the same across each dimension.

We first sample the input data vector $x$ from a standard normal distribution. We randomly sample scalars $a$ and $b$ from the same distribution as the respective true weight and bias parameters of the data. This gives us $y$, the label vector, equal to $ax + b$. Because no realistic data is perfectly linear we add noise to our dataset. We sample noise from a standard normal distribution and then scale the noise to the magnitude of the pre-existing data by multiplying it by the expectation of $y$. This scaled noise is then multiplied by a noise coefficient $\beta$, which controls the extent to which the labels $y$ are corrupted by noise. Finally this noise scaled to the magnitude of the dataset is added to the pre-existing labels $y$ to give the noisy labels, $y_{noise}$. In equation form, our noisy labels are given by: 

\begin{equation}
y_{noise} = ax + b + \beta * \mathcal{N}(0, 1) * \mathbb{E}[ax + b]
\end{equation}

For the new noisy dataset, the new optimal weight is denoted as $a^{*}$ and optimal bias as $b^{*}$.

\subparagraph{Results}
We compare the performances of a linear regression model to LNNs with 2 to 10 layers. For each experiment, using the aforementioned data procedure, we generate a 1000-length data and label vector for model training and a 200-length data and label vector model evaluation. Both datasets are generated with the same noise coefficient. We first train each model on the training data to convergence. At each iteration, we track the model’s MSE on the train and test datasets.

Additionally, we track the model's parameters deviation from the optimal weight and bias at iteration.We calculate the deviation of a given model's parameters from the optimal solution by first applying the Normal Equation, a closed-form solution, on the training data to solve for optimal weight $a^{*}$ and optimal bias $b^{*}$. Because all models are a linear function, we can simplify all models to a linear function $mx + b$ and then measure the model's optimal parameter deviation $D$ as $|m - a^{*}| + |b - b^{*}|$. Over the iterations, this deviation should reduce.

\begin{figure}[H]
\includegraphics[width=7cm]{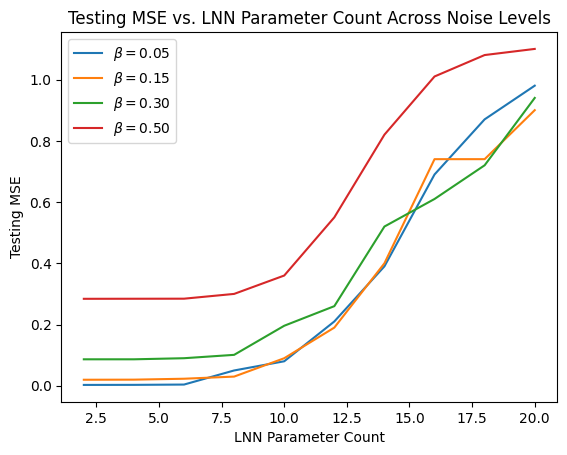}
\centering
\caption{Trendlines of testing MSE as LNN parameter count/layers increases across all noise levels.}
\label{fig:params}
\end{figure}

We perform this experiment 100 times for each of the noise coefficient values 0.05, 0.15, 0.3, and 0.5. We write our models in PyTorch \cite{paszke2019pytorch} and train them with SGD \cite{robbins1951stochastic} using a learning rate of 0.001. We report the testing mean and standard deviations of the MSE (across all 100 experiments) for all models and noise coefficients in Table \ref{table:results}. Figure \ref{fig:D} shows the average optimal parameter deviation $D$ throughout training over the 100 experiments for each model with $\beta=0.05$. Figure \ref{fig:params} shows the sharp increases in MSE as the LNN parameter count (or number of layers) increase across all noise levels.

\hspace{-0.2\textwidth}
\begin{minipage}{.50\textwidth} %
\includegraphics[width=85mm]{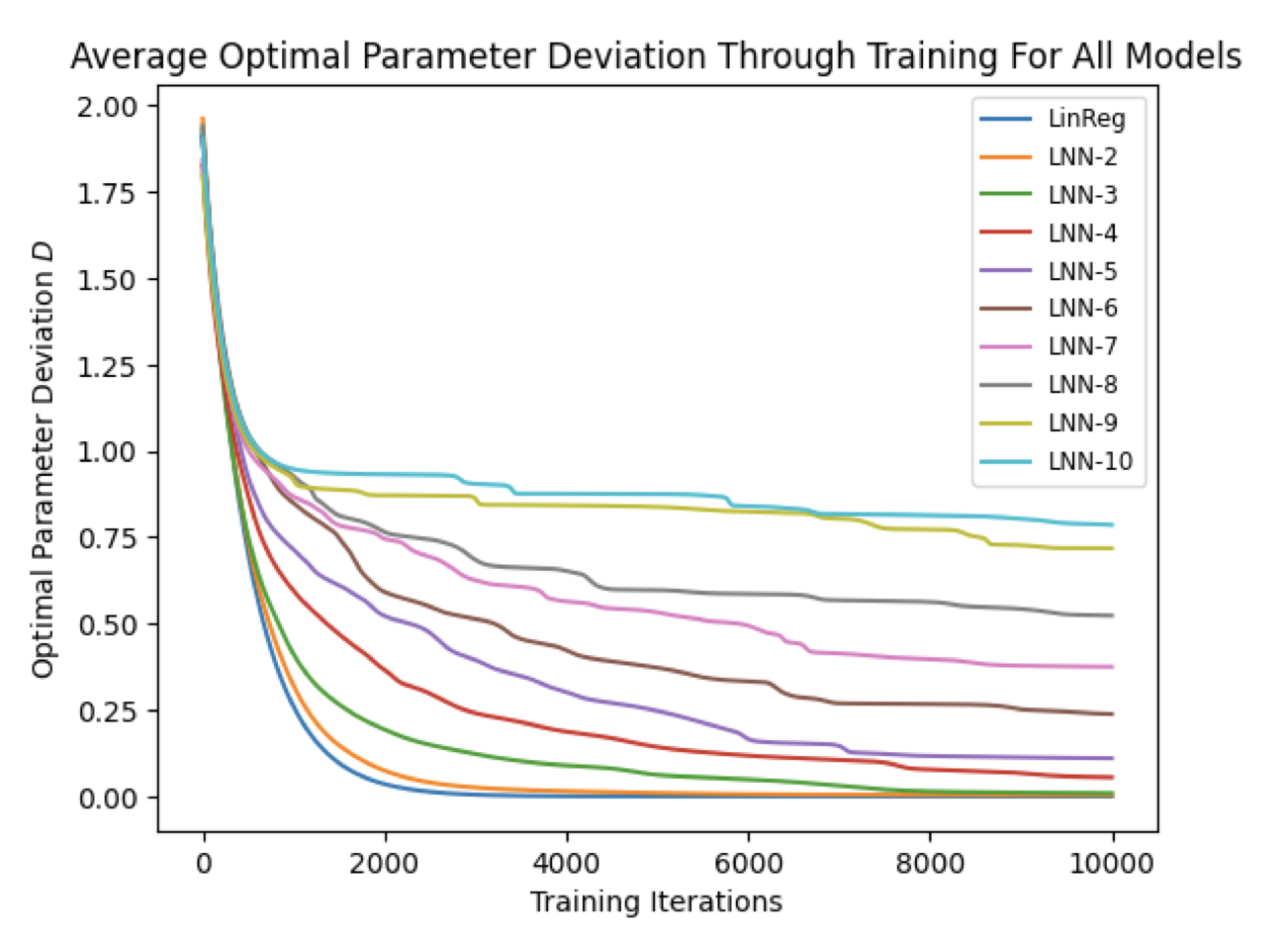}
\captionof{figure}{Plot of the average optimal parameter deviation $D$ for each model across all 100 training runs.}
\label{fig:D}
\end{minipage}
\hspace{0.15\textwidth}
\begin{minipage}{.7\textwidth}%
    \centering
    \renewcommand*{\arraystretch}{1.4}
    \scalebox{0.8}{
    \begin{tabular}{ccccc}
        \toprule
          & \multicolumn{4}{c}{Noise Coefficient $\beta$} \\ 
         Model & 0.05 & 0.15 & 0.30 & 0.50 \\
         \midrule
         LinReg & 0.0028 ±0.005 & 0.0197 ±0.025 & 0.086449 ±0.1197 & 0.2840 ±0.4667
         \\
         LNN-2 & 0.003 ±0.006 & 0.020 ±0.025 & 0.086451 ±0.1197 & 0.2842 ±0.4668 \\
         LNN-3 & 0.004 ±0.007 & 0.023 ±0.04 & 0.09 ±0.1194 & 0.2844 ±0.4665 \\
         LNN-4 & 0.05 ±0.27 & 0.03 ±0.05 & 0.101 ±0.13 & 0.30 ±0.47 \\ 
         LNN-5 & 0.08 ±0.28 & 0.09 ±0.26 & 0.196 ±0.42 & 0.36 ±0.61 \\ 
         LNN-6 & 0.21 ±0.55 & 0.19 ±0.58 & 0.26 ±0.59 & 0.55 ±0.9 \\
         LNN-7 & 0.39 ±0.85 & 0.40 ±0.98 & 0.52 ±1.02 & 0.82 ±1.32 \\ 
         LNN-8 & 0.69 ±1.48 & 0.74 ±1.14 & 0.61 ±0.87 & 1.01 ±1.35 \\ 
         LNN-9 & 0.87 ±1.27 & 0.74 ±1.08 & 0.72 ±1.06 & 1.08 ±1.45 \\ 
         LNN-10 & 0.98 ±1.35 & 0.90 ±1.33 & 0.94 ±1.17 & 1.10 ±1.296 \\ 
         \bottomrule
    \end{tabular}}
    \captionof{table}{Means and standard deviations of testing MSE measured over 100 runs for all models and noise coefficients. LNN-$n$ refers to an LNN with $n$ layers.}
    \label{table:results}
\end{minipage}

\vspace{-0.45cm}


\subparagraph{Discussion} The optimal parameter solution $D=0$ is achieved only by linear regression and LNNs with a few layers. LNNs with more layers typically converge at increasingly suboptimal solutions despite being provided an excessive number of iterations. This highlights the empirical difficulty of excess parameters in  optimization, showing both training and testing performance suffer.

\section{Conclusion}

We are the first to propose a novel explanation against neural networks without activation functions. We  prove the superiority of linear regressions compared to linear neural networks by a comparison of their optimization. We validate this proof by testing linear regression and LNNs on different levels of noise across 100 datasets for each level. We conclude LNNs perform worse in training and tesitng than linear regression due to more difficult optimization caused by their excess parameters.

\printbibliography

@article{robbins1951stochastic,
  title={A stochastic approximation method},
  author={Robbins, Herbert and Monro, Sutton},
  journal={The annals of mathematical statistics},
  pages={400--407},
  year={1951},
  publisher={JSTOR}
}

@article{mcculloch1943logical,
  title={A logical calculus of the ideas immanent in nervous activity},
  author={McCulloch, Warren S and Pitts, Walter},
  journal={The bulletin of mathematical biophysics},
  volume={5},
  pages={115--133},
  year={1943},
  publisher={Springer}
}

@article{paszke2019pytorch,
  title={Pytorch: An imperative style, high-performance deep learning library},
  author={Paszke, Adam and Gross, Sam and Massa, Francisco and Lerer, Adam and Bradbury, James and Chanan, Gregory and Killeen, Trevor and Lin, Zeming and Gimelshein, Natalia and Antiga, Luca and others},
  journal={Advances in neural information processing systems},
  volume={32},
  year={2019}
}

 \end{document}